\pdfoutput=1

\documentclass[11pt]{article}

\usepackage[]{acl}

\usepackage{times}
\usepackage{latexsym}

\usepackage[colorinlistoftodos,prependcaption,textsize=tiny]{todonotes}

\usepackage{soul}

\newcommand{\cmt}[1]{}
\usepackage{hyperref}
\usepackage[ruled,linesnumbered]{algorithm2e}
\usepackage{amsmath}
\usepackage{amsfonts}
\usepackage{amssymb} 
\usepackage{cleveref}
\usepackage{color, colortbl}

\usepackage{multirow}
\usepackage{nccmath}
\usepackage{hhline}
\usepackage{geometry}
\usepackage{soul}
\usepackage{booktabs}
\usepackage[tableposition=above]{caption}
\usepackage{pifont}

\newcommand{\change}[1]{{\textcolor{black}{#1}}}

\definecolor{mypink}{rgb}{0.858, 0.188, 0.478}

\definecolor{yellow}{rgb}{0.9, 0.89, 0.0}
\definecolor{pear}{rgb}{0.82, 0.89, 0.19}
\definecolor{violet}{rgb}{0.5, 0.0, 1.0}
\definecolor{springgreen}{rgb}{0.24, 0.7, 0.44}
\definecolor{orange}{rgb}{1.0, 0.5, 0.0}
\definecolor{blue}{rgb}{0.0, 0.0, 1}
\definecolor{ruddybrown}{rgb}{0.73, 0.4, 0.16}

\usepackage[T1]{fontenc}
\SetKwInput{KwInput}{Input}                
\SetKwInput{KwOutput}{Output}              
\SetKwInput{KwInitialize}{Initialize}              

\usepackage[utf8]{inputenc}
\usepackage{microtype}

\title{Modeling \emph{What-to-ask} and \emph{How-to-ask} for Answer-unaware \\ Conversational Question Generation}

\author{
Xuan Long Do$^{1,2,}$\thanks{\quad Contribution during the internship at Institute for Infocomm Research.}, \ Bowei Zou$^{1}$, \ \textbf{Shafiq Joty}$^{2,3}$\thanks{Work done when the author was on leave from NTU.}\textbf{,} \ Anh Tai Tran$^{4}$,  \\ \textbf{Liangming Pan}$^{5}$\textbf{,} \textbf{Nancy F. Chen}$^{1}$\textbf{,} \ \textbf{Ai Ti Aw}$^{1}$\\
$^1$Institute for Infocomm Research (I2R), A*STAR, \\ $^2$Nanyang Technological University, Singapore, $^3$Salesforce AI,\\ $^4$ByteDance, $^5$University of California, Santa Barbara\\ 
\small{\{doxuanlong15052000, anhtai2672000\}@gmail.com,}\\
\small{liangmingpan@ucsb.edu,} 
\small{\{zou\_bowei, nfychen, aaiti\}@i2r.a-star.edu.sg,} \small{srjoty@ntu.edu.sg} 
}

\begin{document}
\maketitle

\begin{abstract}
Conversational Question Generation (CQG) is a critical task for machines to assist humans {in fulfilling their information needs} through conversations. The task is generally cast into two different settings: answer-aware and answer-unaware. While the former facilitates the models by exposing the expected answer, the latter is more realistic and receiving growing attentions recently. 
\emph{What-to-ask} and \emph{how-to-ask} are the two main challenges in the answer-unaware setting. 
To address the first challenge, existing methods mainly select sequential sentences in context as the rationales.
We argue that the conversation generated using such {naive} heuristics may not be natural enough as in reality,  the interlocutors often talk about the relevant contents that are not necessarily sequential  in context. Additionally, previous methods decide the type of question (boolean/span-based) to be generated  implicitly.
{Modeling the question type explicitly is crucial in this (answer-unaware) setting, as the answer which hints the models to generate a boolean or span-based question, is unavailable.} 
To this end, we present \emph{SG-CQG}, a two-stage CQG framework. For the \emph{what-to-ask} stage, a sentence is selected as the rationale from a semantic graph that we construct, and extract the answer span from it. For the \emph{how-to-ask} stage, a classifier determines the target {answer} type of the question via two explicit control signals before generating and filtering.
In addition, we propose \emph{Conv-Distinct}, a novel evaluation metric for CQG, to evaluate the diversity of the generated conversation from a context.
Compared with the existing answer-unaware CQG models, the proposed \emph{SG-CQG} achieves state-of-the-art performance. 

\end{abstract}

\section{Introduction}


Building systems that can comprehend human speech and provide assistance to humans through conversations is one of the main objectives in AI. Asking questions during a conversation is a crucial conversational behavior that helps AI agents communicate with humans more effectively \cite{James07, li2016learning}. This line of research is known as \emph{Conversational Question Generation (CQG)}, which targets generating questions given the context and conversational history \cite{nakanishi-etal-2019-towards, pan-etal-2019-reinforced, gu-etal-2021-chaincqg, cohs-cqg}. Compared to traditional single-turn question generation \cite{https://doi.org/10.48550/arxiv.1905.08949}, CQG is more challenging as the generated multi-turn questions in a conversation need not only to be coherent but also follow a naturally conversational flow. 

Generally, there are two main settings for the CQG task: answer-aware and answer-unaware. In the answer-aware setting, the expected answers of the (to be) generated questions are exposed to the models \cite{gao-etal-2019-interconnected, gu-etal-2021-chaincqg, shen-etal-2021-gtm, cohs-cqg}. In reality, however, the answers are only ``future'' information that are unknown beforehand. Thus, growing attention has been on the more realistic answer-unaware setting, in which the answers are unknown to the CQG model \cite{wang-etal-2018-learning-ask, pan-etal-2019-reinforced, nakanishi-etal-2019-towards, qi-etal-2020-stay,cohs-cqg}.

Prior studies either attempt to ask the questions first, and compute the reward function to evaluate their answerability \cite{pan-etal-2019-reinforced} or informativeness \cite{qi-etal-2020-stay}; or they extract the answer spans from the context as the \emph{what-to-ask} first, and generate the questions based on them \cite{nakanishi-etal-2019-towards, cohs-cqg}. However, it has been argued that the former approach tends to generate repetitive questions
\cite{qi-etal-2020-stay, cohs-cqg}. 
For the latter approach, \citet{cohs-cqg} recently proposed a selection module to shorten the context and history of the input and achieved state-of-the-art performance. 
Nonetheless, it simply employs a naive heuristic to select the earliest forward sentence (without traceback) in the context as the rationale to extract the answer span. Although such heuristics ensure the flow of the generated questions is aligned with the context, we argue that the resulting conversations may not be natural enough, because, in reality, the interlocutors often talk about the relevant parts that may not form a sequential context. \change{Furthermore, previous studies \cite{gao-etal-2019-interconnected, cohs-cqg} trained the models to decide the type of the question (boolean/span-based) to be generated implicitly. We argue that modeling question type explicitly is critical since in this setting, the answer, which hints the models to generate a boolean or span-based question, is unavailable.}



To address the above problems, we propose a two-stage CQG framework based on a semantic graph, \emph{SG-CQG}, which consists of two main components: \emph{what-to-ask} and \emph{how-to-ask}.
In particular, given the referential context and dialog history, the \emph{what-to-ask} module \emph{(1)} constructs a semantic graph, which integrates the information of coreference, co-occurrence, and named entities from the context to capture the keyword 
chains for the possible ``jumping'' purpose; 
\emph{(2)} 
traverses the graph to retrieve a relevant sentence as the rationale; and \emph{(3)} extracts the expected answer span from the selected rationale (Section~\ref{what-to-ask}). 
Next, the \emph{how-to-ask} module decides the question type (boolean/span-based) via two explicit control signals and conducts question generation and filtering (Section~\ref{how-to-ask}).

In order to exhaustively assess the quality of the generated question-answer pairs, we propose a set of metrics to measure the \emph{diversity}, \emph{dialog entailment}, \emph{relevance}, \emph{flexibility}, and \emph{context coverage} through both standard and human evaluations. 
Compared with the existing answer-unaware CQG models, our proposed \emph{SG-CQG} achieves state-of-the-art performance on the standard benchmark, namely the CoQA dataset \cite{reddy-etal-2019-coqa}. 

Our contributions can be summarized as follows:

\emph{(1)} We propose \emph{SG-CQG}, a two-stage framework, which consists of two novel modules: \emph{what-to-ask} encourages the models to generate coherent conversations; and \emph{how-to-ask} promotes generating naturally diverse questions. Our codes will be released at \url{https://github.com/dxlong2000/SG-CQG}.

\emph{(2)} \emph{SG-CQG} achieves state-of-the-art performance on answer-unaware CQG on CoQA. 

\emph{(3)} To the best of our knowledge, we are the first to propose a set of criteria to comprehensively evaluate the generated conversations. Moreover, we propose \emph{Conv-Distinct} to measure the diversity of the generated conversation from a context, which takes the context coverage into account.

\emph{(4)} We conduct thorough analysis and evaluation of the questions and answers of our generated conversations, which can bring some inspiration for future work on the answer-unaware CQG. 

\section{Related Work}
Our work is closely related to two lines of prior work. Extended related work is in \Cref{appendix:extended-related-work}.
\subsection{Conversational Question Generation}
Question Generation has gained much attention from the research community over the years \cite{https://doi.org/10.48550/arxiv.1905.08949, lu-lu-2021-survey}. Despite such intensive exploration, much less attention has been drawn to {Conversational QG or CQG}. Generally, CQG has been considered in two main settings: answer-aware and answer-unaware. In the answer-aware setting, the expected answers are revealed to models \cite{gao-etal-2019-interconnected, gu-etal-2021-chaincqg, shen-etal-2021-gtm, cohs-cqg}. However, this is not always the case in reality, as the answers are ``future information''. The answer-unaware setting; therefore, receives growing interests recently \cite{wang-etal-2018-learning-ask, pan-etal-2019-reinforced, nakanishi-etal-2019-towards, qi-etal-2020-stay,cohs-cqg}. 

To tackle the \emph{what-to-ask} problem, prior studies \cite{pan-etal-2019-reinforced, cohs-cqg} selected the next sentence in the context as the rationale. \citet{cohs-cqg} extract the target answer span from the rationale, while \citet{pan-etal-2019-reinforced} generate the question, and compute a reward function to fine-tune the model by reinforcement learning. The \emph{how-to-ask} challenge was \change{simply} formulated as that in the answer-aware setting. 
In contrast, we attempt to model the rationale selection in a more coherent way by constructing and traversing a semantic graph, which simulates the keyword chains. We further propose control signals to promote diversity and fluency in question generation.


\subsection{Knowledge-grounded Conversation Generation}
Leveraging graphs to enhance dialog response generation has received growing interest \cite{moghe-etal-2018-towards, liu-etal-2019-knowledge, ijcai2020p545, xu-etal-2021-discovering}. In particular, \citet{ijcai2020p545} proposed to extract event chains  \cite{mostafazadeh-etal-2016-corpus}, and utilised them to help determine a sketch of a multi-turn dialog. Nonetheless, the situation differs significantly when it comes to the CQG task. The responses in the dialog response generation task are normally full sentences with enough relevant mentions. However, in CQG, the questions and answers are mostly short and lack clear keywords, which makes the \change{existing} 
keyword-graph not applicable. We thus present a semantic graph, which incorporates the coreference, co-occurrence, and named entities information from the context.

\section{\emph{SG-CQG}} 
\begin{figure*}[t!]
     \centering
        \includegraphics[trim={0cm 0cm 0cm 0cm},clip]{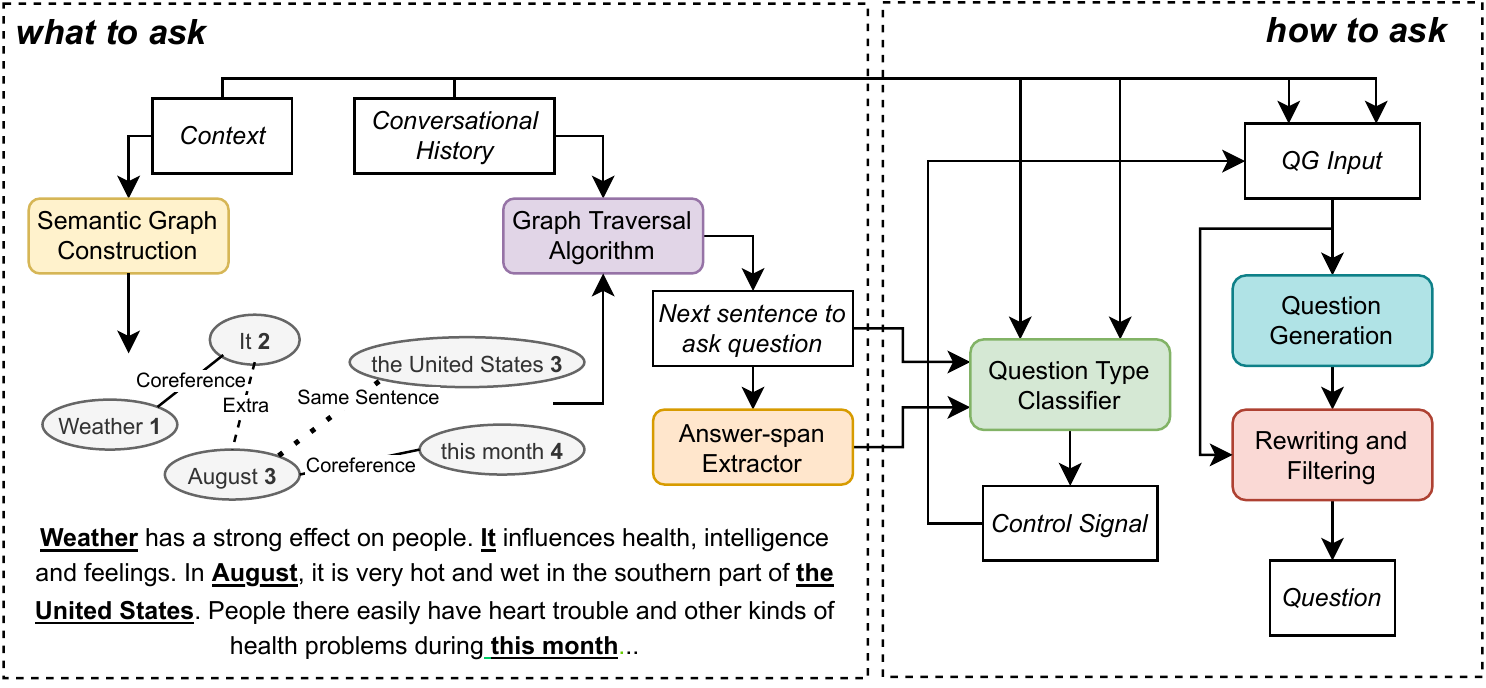}
        \vspace{-3mm}
         \caption{\small{An overview of our proposed \emph{SG-CQG} framework. It consists of two modules: the \emph{what-to-ask} module aims to select a sentence as the rationale from the context and extracts the target answer span from it, and the \emph{how-to-ask} module then predicts the type of the question to be generated and generates the question guided by that type.}
         }
    \label{fiq:overview-system}
\end{figure*} 

We formulate the answer-unaware conversational question generation (CQG) task as: given the referential context $C = \{s_1, s_2, ..., s_m\}$ with  $s_i$ being the $i$-th sentence in context, and the conversational history $H_n=\{(q_1, a_1), (q_2, a_2), ..., (q_{n-1}, a_{n-1})\}$ with $(q_i, a_i)$ being the $i$-th turn of the question-answer pairs, as input $\mathcal{D}_n = \{C, H_n\}$, the model learns to generate the current question $q_n$ and answer $a_n$. 

\Cref{fiq:overview-system} demonstrates an overview of our proposed framework. It consists of two main components: \emph{(1)} A \emph{what-to-ask} module aims to select a reasonable sentence in the referential context $C$ as the current rationale $r_n$ and thereby a span in $r_n$ as the target answer $a_n$, given $\mathcal{D}_n$. \emph{(2)} A \emph{how-to-ask} module aims to generate the question $q_n$, guided by the rationale $r_n$ and target answer $a_n$.

\subsection{\emph{What-to-ask} Module (WTA)} \label{what-to-ask}
Existing answer-unaware CQG models \cite{pan-etal-2019-reinforced, cohs-cqg} commonly utilize the next sentence of $r_{n-1}$ in the context as the current rationale $r_n$. Although such heuristics can guarantee that the flow of the generated questions is consistent with the narrative in context, the generated conversation may not always be as natural as in reality, since human speakers often jump back and forth across the relevant but not sequential contents in context. To facilitate the models in selecting the current rationale and target answer appropriately and further improve the semantic diversity of dialogue flow, we design a \emph{what-to-ask} module, which consists of two components: \emph{semantic graph construction} and \emph{graph traversal algorithm}. 

\paragraph{Semantic Graph Construction (SGC)} \Cref{fiq:overview-system} shows an example of our semantic graph. Each node is displayed as a textual span and the index of the sentence it belongs to. To construct the semantic graph $\mathcal{G} = \{\mathcal{V}, \mathcal{E}\}$, we first obtain the coreference clusters from the context $C$ by AllenNLP \cite{https://doi.org/10.48550/arxiv.1904.05255} and build the set of initial nodes from phrases in the clusters. We then connect all the nodes in the same cluster as a chain: each node in the cluster (except the one that appears last in the context) is connected to the nearest forward one in the context. We denote this type of relation as \emph{Coreference}. To enhance the connectedness of $\mathcal{G}$, we extract all named entities by \emph{spaCy}\footnote{https://spacy.io/} and add them as additional nodes if they are not in any clusters. We then connect all the nodes in the same sentence in the context in the same chaining style and name those edges as \emph{Same Sentence}. 
Finally, we add a type of \emph{Extra} edges between all connected subgraphs to make $\mathcal{G}$ fully-connected. Since those \emph{Extra} edges do not bring any semantic relation to the graph, our objective is to minimize the number of those edges. Specifically, we gradually select, and connect two sentences such that their nodes are in different connected components and have the smallest indexes with the smallest difference, until the graph is fully-connected. To connect two sentences, we add an \emph{Extra} edge between the last phrase in the smaller-index sentence and the first phrase in the remaining sentence. The adding-\emph{Extra}-edges algorithm is in \Cref{appendix:component-merging-algorithm}. 

\paragraph{Graph Traversal Algorithm (GTA)} 
Given the conversational history $H_n$ and the semantic graph $\mathcal{G}$, we create a queue $q$ to store nodes for traversing. We first add the nodes that appear in any previous turn' rationale to $q$ in the index order \footnote{The nodes are ordered according to their sentences' indexes in the original context.}. 
We then traverse $\mathcal{G}$ by popping the nodes in $q$ until it becomes empty. For each node, we retrieve the sentence that contains it as the rationale $r_n$. If the model can generate a valid question from $r_n$ and any answer span extracted from $r_n$, we add all unvisited neighbors of the current node to the beginning of $q$. A question is considered being valid if it passes the QF module (\Cref{ssub-rewriting-and-filtering}). Prepending the neighbors to queue is to prioritize the nodes that are connected so that the generated conversation can be formed from a chain of relevant sentences, which consolidates the coherence of the conversation. If the model cannot generate any valid $q_n$ by the current node, we add its unvisited neighbors to the end of $q$. The pseudocode of our proposed \emph{Graph Traversal Algorithm} is described in \Cref{appendix:graph-traversal-algorithm}. 

\paragraph{Answer Span Extractor (AE)} We follow \citet{cohs-cqg} to design the answer span extractor module. In particular, a T5 model is trained on SQuAD \cite{rajpurkar-etal-2016-squad} 
to predict the target answer span ($a$), given its original sentence in context ($r$). We use this pretrained model to extract $a_n$ from $r_n$. Note that we also deselect the answer spans that are the same as those of previous turns. 

\subsection{\emph{How-to-ask} Module (HTA)} \label{how-to-ask}
A high ratio of boolean questions in conversational datasets such as CoQA \cite{reddy-etal-2019-coqa} (around 20\%) is one of the main challenges for current CQG studies \cite{gao-etal-2019-interconnected, pan-etal-2019-reinforced, gu-etal-2021-chaincqg}. To the best of our knowledge; however, there is no up-to-date work which attempts to tackle this challenge. This problem is even worse in the answer-unaware setting since there is no \emph{\texttt{Yes/No}} answer to be provided to guide the generation of the models. Previous studies \cite{pan-etal-2019-reinforced, cohs-cqg} simply train the CQG models to let them implicitly decide when to generate the boolean and span-based questions without any explicit modeling of the question type. We argue that explicitly modeling the question type is critical, as the models will gain more control on generating diverse questions, thus making the conversation become more natural. To this end, we introduce two control signals as the additional input to the QG model, and develop a simple mechanism to select the signal for the current turn.

\paragraph{Question Type Classifier (QTC)} We design two control signals to guide the QG model: \texttt{<BOOLEAN>} is prepended to the textual input if we expect the model to generate a boolean question, and \texttt{<NORMAL>} otherwise. To classify which signal should be sent to the QG model, we train a RoBERTa \cite{roberta-paper} as our \emph{Question Type Classifier}. This binary clasifier takes the rationale $r_n$ and the answer span $a_n$ generated from \emph{what-to-ask} module, the context and the shortened conversational history as the input, and generates the label $0/1$ corresponding to \texttt{<NORMAL>/<BOOLEAN>}. We conduct additional experiments to discuss why the $control\_signals$ work in \Cref{ssec:how-do-control-signals-work}.

\begin{table}[t!]
\centering
\scalebox{0.7}
{\begin{tabular}{p{2.5cm}p{7cm}}
\toprule
\textbf{Type}   & \textbf{Example} \\
\midrule
Wrong answer & 'Did he eat for \textcolor{red}{breakfast}?', '\textcolor{red}{breakfast}'\\
Irrelevant & 'Was he still \textcolor{red}{alive}?', 'no', \\
Uninformative & 'What happened one day?', ‘\textcolor{red}{Justin} woke up very excited’, '\textcolor{red}{Who woke up?}', ‘\textcolor{red}{Justine}’\\
Redundant & '\textcolor{red}{Did he eat something?}', 'yes',..., '\textcolor{red}{Was he eating something?}', 'yes'\\
\bottomrule
\end{tabular}}
\vspace{1mm}
\caption{\small Different types of common errors that CQG models are prone to without our extra postprocessing heuristics.}
\vspace{-5mm}
\label{tab:question-error}
\end{table}

\paragraph{Rewriting and Filtering (RF)}  \label{ssub-rewriting-and-filtering}
Our RF module serves two purposes. Firstly, following \citet{cohs-cqg}, we train a T5 model on CoQA \cite{reddy-etal-2019-coqa} as our CQA model to answer the generated questions. A question is passed this filtering step if the answer generated by the CQA model has a fuzzy matching score greater or equal to 0.8 with the input answer span. Secondly, when invigilating the generated conversations, we observe multiple other errors that the blackbox model encounters, as shown in \Cref{tab:question-error}. We thus propose extra post-processing heuristics to filter out the generated questions and try to avoid the following issues: \emph{(1)} \emph{Wrong answer}. Unlike \citet{cohs-cqg} that took the extracted spans as the conversational answers, 
we rewrite the extracted answer spans for the boolean questions by selecting the answers generated from the CQA model; \emph{(2)} \emph{Irrelevant}. For each generated question, we remove stopwords and question marks only for filtering purpose, and we check if all the remaining tokens exist in the context $C$; 
\emph{(3)} \emph{Uninformative}. To remove the turns like \textit{(``Who woke up?'', ``Justine'')}, we check validity if no more than 50\% of the tokens of $r_n$ exist in any previously generated QA pairs; \emph{(4)} \emph{Redundant}. Unlike previous studies \cite{qi-etal-2020-stay, cohs-cqg} which only considered the redundant information from the generated answers, for each generated question that has more than 3 tokens, we filter it out if it has a fuzzy matching score >= 0.8 with any of the previously generated questions.

\paragraph{Question Generation (QG)} 
We fine-tune a T5 model \cite{JMLR:v21:20-074} to generate conversational questions. We concatenate the input $\mathcal{D}^a_n = \{C, H_n, a_n, r_n, control\_signal\}$ in the format: \texttt{Signal}: $control\_signal$ \texttt{Answer}: \textit{$a_n$}, \textit{$r_n$} \texttt{Context:} \textit{$C$} \texttt{[SEP]} $H_{sub}$, where $H_{sub} \in H_n$. The model then learns to generate the target question $q_n$. In our experiments, $H_{sub}$ is the shortened $H_n$, in which we keep at most three previous turns. It was shown to improve upon training with the whole $H_n$ significantly \cite{cohs-cqg}. The performance of the QG model is in \Cref{appendix:question-generation-performance}.

\section{Experimentation} \label{sec:system-experiment}
\subsection{Experimental Settings}

\begin{table*}[htbp]
\centering
\scalebox{0.8}{
\makebox[\textwidth][c]{
    \begin{tabular}{l|cc|cc|ccc|c|c|c}
\toprule
         & \multicolumn{2}{c|}{Distinct} & 
         \multicolumn{2}{c|}{\emph{Conv-Distinct}} &
         \multicolumn{3}{c|}{BERTScore-entailment} &
         \multicolumn{1}{c|}{BERTScore} &  
         \multicolumn{1}{c|}{\emph{CC (\%)}} & \multicolumn{1}{c}{\emph{JS (\%)}} \\
\midrule
Model & 1  & 2 & 1 & 2 & 1 & 2 & all & & & \\ 
\hline
BART & \bf{84.09} & \bf{97.25} & 6.89 & 8.28 & 48.77 & 48.83 & 48.76 & \bf{82.07} & 8.62 & 15.07 \\
T5 & 60.31 & 82.20 & 14.44 & 19.77 & 77.51 & 79.23 & 77.01 & 81.13 & 23.33 & 13.83 \\
GPT-2 & 60.12 & 88.06 & 19.72 & 26.99 & 77.77 & 79.70 & 77.18 & 79.49 & 34.50 & 7.50 \\
CoHS-CQG & 67.17 & 92.65 & 20.11 & 27.52 & 77.97 & 79.24 & 77.62 & 80.79 & 30.02 & 0.00\\
\hline
\emph{SG-CQG $+$ w/o WTA} & 72.13 & 95.21 & 20.95 & 27.78 & 77.63 & 79.35 & 78.02 & 80.79 & 29.21 & 0.00 \\
\emph{SG-CQG $+$ w/o RF} & 21.00 & 50.01 & 21.00 & 50.01 & 80.55 & 81.16 & 78.13 & 77.74 & \underline{100.00} & 6.69 \\
\emph{SG-CQG $+$ w/o QTC} & 57.47 & 91.28 & 38.93 & 62.13 & 81.95 & 83.20 & 79.18 & 80.76 & 68.06 & 19.67 \\
\emph{SG-CQG (ours)} & 57.42 & 91.29 & \bf{38.99{$\dagger$}} & \bf{62.27{$\dagger$}} & \bf{81.99{$\dagger$}} & \bf{83.27{$\dagger$}} & \bf{79.29{$\dagger$}} & 80.89 & \bf{68.52$\dagger$} & 19.72\\
\hline
\hline
\emph{Oracle} & 58.29 & 80.10 & 33.60 & 52.89 & 81.93 & 82.95 & 79.36 & 81.05 & 58.10 & 16.11 \\
\bottomrule
\end{tabular}
}}

\vspace{1mm}
\caption{\small{Performance of answer-unaware CQG on the test set (CoQA dev set). CC: Context Coverage score, JS: Jumping Score. \change{$\dagger$ denotes our model significantly outperforms baselines with p-value < 0.01 under t-test (\Cref{appendix:statistical-significance}).}}}

\label{tab:main-evaluation}
\vspace{-5mm}
\end{table*}

\begin{table}[t!]
\begin{center}
\resizebox{.65\columnwidth}{!}{%
\makebox[0.45\textwidth][c]{
    \begin{tabular}{l|cc|cc|c}
\toprule
         & \multicolumn{2}{c|}{Distinct} & 
         \multicolumn{2}{c|}{\emph{Conv-Distinct}} &
         \multicolumn{1}{c}{CC (\%)}\\
\midrule
Model & 1  & 2 & 1 & 2 \\ 
\hline
ReDR & 22.15 & 33.42 & - & - & -  \\
T5 & 51.17 & 73.07 & 12.98 & 17.58 & 23.33 \\
GPT-2 & 57.79 & 88.04 & 18.89 & 24.93 & 34.50 \\
CoHS-CQG & 66.18 & 90.01 & 19.05 & 25.67 & 30.02\\
\hline
\emph{SG-CQG $+$ w/o WTA} & \bf{68.35} & \bf{92.33} & 19.66 & 26.47 & 29.21 \\
\emph{SG-CQG $+$ w/o RF} & 23.48 & 51.14 & 23.48 & 51.14 & \underline{100.00} \\
\emph{SG-CQG $+$ w/o QTC} & 49.27 & 79.53 & 33.18 & 54.04 & 68.06 \\
\emph{SG-CQG} & 54.15 & 79.61 & \bf{33.34} & \bf{54.26} & \bf{68.52} \\
\hline
\hline
Oracle & 54.91 & 85.76 & 31.87 & 49.86 & 58.10 \\
\bottomrule
\end{tabular}%
}
}
\end{center}
\caption{\small{Question generation evaluation results on our test set (CoQA validation set).}}

\label{tab:question-evaluation}
\vspace{-5mm}
\end{table}

\paragraph{Dataset}
We use CoQA \cite{reddy-etal-2019-coqa}, a large-scale CQA dataset, in our experiments. Each conversation includes a referential context and multiple question-answer pairs, resulting in a total of 127k question-answer pairs. Among them, around 20\% of questions are boolean, which makes this dataset become challenging for the CQG task \cite{pan-etal-2019-reinforced, gu-etal-2021-chaincqg}. Since the test set of CoQA is unavailable, we follow \citet{cohs-cqg} to keep the original validation set as our \emph{test set} and randomly sample 10\% of the original training set as our new \emph{validation set}.

\paragraph{Automatic Evaluation} 
We utilise BERTScore \cite{Zhang2020BERTScore:} as our dialog entailment metric \change{(BERTScore-entailment)}, a generalization of \citet{dziri-etal-2019-evaluating-coherence}. It considers the generated response (question/answer) as the premise, and the utterances in the conversational history as the hypothesis, and measures their similarity score as the topic coherence score. \change{This property is crucial as the questions/answers should focus on the same topic as the previous turn(s).} In our experiment, we measure the dialog entailment score with 1, 2, and all previous turn(s). To measure the relevance between the generated conversation and the context, we concatenate the generated QA pairs and compute the BERTScore. \change{It provides how the generated conversation is explicitly relevant to the context.}

We observe short conversations with very few generated turns tend to yield very high scores on the available diversity measurement metrics such as Distinct \cite{li-etal-2016-diversity}. Since the conversation is generated from a given context, we argue that how much information from the given context the generated conversation covers should be taken into account. To this end, we introduce \emph{Context Coverage (CC)} to measure the percentage of the sentences in the context that are the rationales of generated QA pairs. Our proposed \emph{Conv-Distinct} of a generated conversation is then computed by multiplying the Distinct score of the generated conversation with its CC score, to measure the diversity of the turns generated \emph{from a given context}:

\begin{ceqn}    
    \begin{align}
         \label{eq:p1-c1} & \resizebox{.6\hsize}{!}{\emph{Conv-Distinct} =  \emph{CC} * Distinct} 
    \end{align}
\end{ceqn}

\change{We further provide \emph{Jumping Score} (JS) to measure the flexibility of the generated conversation. JS is defined as the percentage of turns in which the model jumps back to any previous content of their previous turn (i.e. trace-back). It is worth noting that we do not rank the models based on JS score. Details of proposed metrics are in \Cref{appendix:evaluation-metric-calculation}.}

\paragraph{Human Evaluation}
Human evaluation is critical to evaluate the quality of the generated conversations since the CQG model may generate reasonable conversations but unmatched well with the provided ground-truth ones. 
We randomly select 25 contexts in our test set and take the first five generated turns from the output of each model to compare, resulting in 125 samples in total. We hire three annotators who are English native speakers. Each generated question is rated by annotators on a 1-3 scale (3 is the best). We follow \citet{cohs-cqg} to utilize three criteria: \textbf{(1) Factuality} measures the factual correctness and meaning of generated questions, \textbf{(2) Conversational Alignment} measures how aligned the generated questions are with the history, \textbf{(3) Answerability} measures how answerable the generated questions are by the given context. Given the fact that LMs can generate fluent texts, we omit using \emph{Fluency} and \emph{Grammaticality}. We measure the annotators' agreement by Krippendorff’s alpha \cite{krippendorff2011computing}. Our human rating instructions are in \Cref{appendix:human-rating-system}.

\paragraph{Implementation Details} 
We fine-tune a RoBERTa$_{large}$ \cite{roberta-paper} as our binary \emph{Question Type Classifier} with the pretrained checkpoints from fairseq \cite{ott2019fairseq} on CoQA. We use a learning rate of 1e-5, a window size of 512, a batch size of 4, and AdamW \cite{loshchilov2018decoupled} as our optimizer. Our classifier achieves an accuracy of 95.6\%. The model is finetuned on a P40 Colab GPU for 10 epochs. Details of the input format are in \Cref{appendix:question-type-classifier-input}.

We initialise \emph{SG-CQG} with pretrained checkpoints of T5$_{base}$ model \cite{JMLR:v21:20-074} from Huggingface \cite{wolf-etal-2020-transformers}. We also use AdamW \cite{loshchilov2018decoupled} as our optimizer with a warmup of 0.1 and an initial learning rate of 1e-4. We train the model for 100k iterations with a standard window size of 512, a batch size of 4, and use a Beam search decoding strategy with a beam size of 4.

\section{Main Results} \label{sec:experimental-results}

\begin{table}{
\begin{center}
\resizebox{.9\columnwidth}{!}{%
\begin{tabular}{l|c|c|c}
    \hline
    Model & EM (\%) & F1 (\%) & CC (\%) \\
    \hline
     GPT-2 & 17.28 & 30.22 & 34.50 \\
     BART  & 18.64 & 38.23 & 8.62 \\
     T5  & 34.29 & 48.67 & 23.33 \\
     CoHS-CQG & 35.14 & 52.08 & 30.02 \\
    \hline
    \emph{SG-CQG $+$ w/o WTA} & 38.89 & 56.17 & 29.21 \\
    \emph{SG-CQG $+$ w/o RF} & 18.14 & 22.85 & \underline{100.00} \\
    \emph{SG-CQG $+$ w/o QTC} & 37.43 & 56.83 & 68.06 \\
    \emph{SG-CQG} & \bf{42.89} & \bf{63.48} & \bf{68.52} \\
    \hline
    \hline
    Oracle & 63.65 & 74.08 & 58.10 \\
    \hline
    \end{tabular}%
}
\end{center}
}
\caption{\small{Answer span extraction evaluation results on our test set (CoQA validation set).}}
\label{tab:answer-evaluation}
\vspace{-9mm}
\end{table}

To evaluate the performance of SG-CQG on the answer-unaware CQG task, we employ 4 baselines for comparison, as shown in \Cref{tab:main-evaluation}. \emph{(1)} T5$_{base}$ \cite{JMLR:v21:20-074}, \emph{(2)} BART$_{base}$ \cite{lewis-etal-2020-bart}, \emph{(3) GPT-2 \cite{radford2019language}}, which are fine-tuned to generate conversational question-answer pairs end-to-end, and \emph{(4)} CoHS-CQG \cite{cohs-cqg} which adopts a strategy to shorten the context and history of the input, achieves the SoTA performance on CoQA in answer-aware and answer-unaware CQG. 

Firstly, we observe that SG-CQG outperforms other methods on most of the metrics, except Distinct and BERTScore. The reason is that BART and T5 often generate short QA pairs (the CC scores are $8.62\%$ and $23.33\%$ on average, respectively),  and copy more from the context, thus they get higher scores on Distinct and BERTScore. Secondly, the metric Conv-Distinct reasonably penalizes models that generate too short conversations, on which SG-CQG achieves the best results. Thirdly, by allowing the model to jump back and forth across the relevant contents in the context by the semantic graph, SG-CQG outperforms other methods significantly on BERTScore-entailment, which indicates that conversational coherence is indeed improved. \change{Furthermore, SG-CQG achieves the highest JS score, which demonstrates that the \emph{what-to-ask} module allows our model to be most flexible in selecting rationales compared to the baselines}. SG-CQG also achieves a significantly higher Context Coverage (CC) score compared to CoHS-CQG. Finally, compared with the results of Oracle, which are from the human-generated conversations, SG-CQG achieves commensurate performance on BERTScore-entailment and BERTScore. It demonstrates that our generated conversations are as closely coherent as human-generated ones. 

\paragraph{Question Generation Evaluation} 
We compare the generated conversational questions of our model with 4 baselines: \emph{(1)} ReDR \cite{pan-etal-2019-reinforced} is an encoder-decoder framework which incorporates a reasoning procedure to better understand what has been asked and what to ask next about the passage; \emph{(2)} T5$_{base}$ \cite{JMLR:v21:20-074}; \emph{(3) GPT-2 \cite{radford2019language}}; \emph{(4)} CoHS-CQG \cite{cohs-cqg}. For T5, GPT-2 and  CoHS-CQG, we extract the generated questions from the generated conversations for comparison. We measure the diversity of the generated questions by Distinct \cite{li-etal-2016-diversity} and our proposed Conv-Distinct. \Cref{tab:question-evaluation} shows evaluation results of the generated conversational questions. We observe that \emph{SG-CQG} achieves the best performance on Conv-Distinct, which takes the context coverage into account.

\paragraph{Answer Span Extraction Evaluation}
We further evaluate the generated conversational answers of our model with 4 baselines: \emph{(1)} T5$_{base}$ \cite{JMLR:v21:20-074}; \emph{(2)} BART$_{base}$ \cite{lewis-etal-2020-bart}; \emph{(3)} GPT-2 \cite{radford2019language}; \emph{(4)} CoHS-CQG \cite{cohs-cqg}. We extract the generated conversational answers from the generated conversations of the models for comparison. We train another T5$_{base}$ model on CoQA for the CQA task (see \Cref{appendix:details-cqa-model}) and utilize it to generate the \emph{ground-truth} answers for the generated questions of the models. We then evaluate the quality of the generated conversational answers by measuring the \emph{Exact Match (EM)} and \emph{F1} scores with the \emph{ground-truth} ones. \Cref{tab:answer-evaluation} shows the evaluation results. We observe that the generated conversational answers extracted by \emph{SG-CQG} achieve the best EM and F1 scores, which are significantly higher than the other baselines.

\paragraph{Human Evaluation} 

\definecolor{gray}{gray}{0.75}
\begin{table}[t!]
\begin{center}
\resizebox{0.4\textwidth}{!}{%
\begin{tabular}{l|c|c|c}
    \hline
    Model & Fact. & C-Align & Ans. \\
    \hline
     T5  & 2.53 & 2.49 & 2.39  \\
     CoHS-CQG & 2.54 & 2.52 & 2.46  \\
    \hline
    \emph{SG-CQG} & \textbf{2.61} & \textbf{2.62} & \textbf{2.53} \\
    \hline
    \hline
     Krip.'s $\alpha$ & 0.71 & 0.72 & 0.75 \\
    \hline
    \end{tabular}
}
\end{center}
\caption{\small{Human evaluation results. \textit{Fact.}: Factuality, \textit{C-Align}: Conversational Alignment, \textit{Ans.}: Answerability.}}
\label{tab:human-evaluation}
\vspace{-5mm}
\end{table}

The results of the human evaluation are present in \Cref{tab:human-evaluation}. Generally, \emph{SG-CQG} achieves the highest performances on all three proposed metrics with a good overall annotators' agreement with an alpha of 0.73. In particular, we observe that by integrating the semantic graph into the selection of the rationales, \emph{SG-CQG} outperforms CoHS-CQG \cite{cohs-cqg} significantly in the conversational alignment property. Furthermore, \emph{SG-CQG} improves CoHS-CQG by a gap in the answerability and factuality of the generated questions, which reflects that our RF module with additional post-processing steps works as expected. 

\section{Discussion} \label{sec:discussion}
\subsection{Ablation Studies} \label{ssec:ablation-studies}

\begin{table*}[t!]
\centering
\small 
\resizebox{1\linewidth}{!}
{
\begin{tabular}{p{7cm}p{6cm}p{1.5cm}}
\toprule
\textbf{Context} & \textbf{Generated Conversation} & \textbf{Rationales} \\
\midrule
1. One day Mary \textcolor{orange}{took a walk to the park}.\newline 
2. The park was very close to her house. \newline 
3. On her way to the park she passed her friend \textcolor{violet}{Kim's house}. \newline 
... \newline 
7. John's house was \textcolor{springgreen}{three houses} down. \newline 
8. Mary and Kim stopped by to ask \textcolor{blue}{John} if he wanted to play at the park. \newline 
... \newline 
14. They loved the \textcolor{yellow}{flowers} and the swings! \newline 
15. Soon it was \textcolor{ruddybrown}{dinnertime} and the girls went home. \newline 
&
Q1: What did Mary do? \newline 
A1: \textcolor{orange}{Took a walk to the park} \newline 
Q2: Where did she see her friend? \newline 
A2: \textcolor{violet}{Kim's house} \newline 
Q3: Who did they ask about going there? \newline 
A3: \textcolor{blue}{John} \newline  
Q4: How far away was his home? \newline 
A4:  \textcolor{springgreen}{Three houses} \newline 
Q5: What time of day were they leaving? \newline 
A5: \textcolor{ruddybrown}{Dinnertime} \newline 
Q6: Did they enjoy \textcolor{yellow}{flowers}? A6: Yes \newline 
&   
1,\newline 3,\newline 8,\newline 7,\newline 15,\newline 14 \\[-0.5em]
\bottomrule
\end{tabular}
}
\vspace{0mm}
\caption{\small{One sample conversation generated by our model SG-CQG.}
}
\vspace{-5mm}
\label{tab:case-studies}
\end{table*}

\paragraph{Ablation of What-to-ask Module (WTA)} \label{sec:ablation-what-to-ask}
To better understand how the \emph{what-to-ask} module affects our proposed model in generating conversations, we study its ablation named \emph{SG-CQG + w/o WTA} in Tables \ref{tab:main-evaluation}, \ref{tab:question-evaluation}, \ref{tab:answer-evaluation}. In this case, our model becomes an upgraded version of CoHS-CQG \cite{cohs-cqg}. Compared to CoHS-CQG, it achieves higher scores on all metrics except the Context Coverage (CC), which reflects that the quality of the generated conversations is indeed improved. These improvements are expected as the model in this case gains more control over generating boolean questions and has a stricter filtering process. This stricter filtering process also explains why it gets a lower CC score compared to CoHS-CQG.

\paragraph{Ablation of Question Type Classifier (QTC)} \label{sec:ablation-question-type-classifier}
We conduct an ablation study of the Question Type Classifier (QTC) module. We name this experiment \emph{SG-CQG + w/o QTC}. \Cref{tab:main-evaluation} shows the evaluation results of generated question-answer pairs. Compared with SG-CQG, the performance of \emph{SG-CQG + w/o QTC} drops slightly on nearly all metrics (except Distinct), which consolidates our hypothesis that explicitly modeling the question type improves the overall coherency of the conversation. 
Furthermore, \Cref{tab:question-evaluation} shows that QTC enhances the diversity of the generated questions, while \Cref{tab:answer-evaluation} illustrates that QTC improves the quality of the generated answers. 

\paragraph{Ablation of Rewriting and Filtering (RF)} \label{sec:ablation-question-filtering}
\emph{SG-CQG + w/o RF} in \Cref{tab:main-evaluation} shows the ablation results of the Rewriting and Filtering (RF) module.
As removing the RF module means we do not filter out any generated question, it results in two consequences. Firstly, since for each sentence, the model can generate at least one conversational question, the CC score of \emph{SG-CQG + w/o RF} is perfect (100\%). Second, redundant questions and answers are generated very frequently. As such, removing the RF module reduces the quality of the generated question-answer pairs (\Cref{tab:main-evaluation}) and questions (\Cref{tab:question-evaluation}) significantly. Notably, without the RF module, the extracted answer spans by \emph{SG-CQG + w/o RF} can be very different from the true conversational answers, resulting in very low F1 and EM scores (\Cref{tab:answer-evaluation}). Although the CC score is perfect, the generated question-answer pairs from this experiment are of bad-quality. 

\subsection{Case Study} \label{ssec:case-study}
We present one conversation generated by our proposed SG-CQG in \Cref{tab:case-studies}. We observe that the rationale of Q2-A2 is the $3$-{rd} sentence in the context, and the rationale of Q3-A3 is the $8$-{th} sentence, which is a forward jump of the model. On the other hand, the rationale of the Q4-A4 is the $7$-{th} sentence, which is a traceback. Such a traceback enhances reasonable coherence between Q3-A3 and Q4-A4. Furthermore, Q5-A5 to Q6-A6 is also a traceback, and especially, Q6 is a boolean question. More case studies are shown in \Cref{appendix:extended-case-studies}.

\subsection{Why Do Control Signals Work?} \label{ssec:how-do-control-signals-work}
\paragraph{Experimental Settings} 
We design the experiments to verify the helpfulness of our two proposed \emph{control\_signals}: \texttt{<BOOLEAN>} and \texttt{<NORMAL>}. In particular, we train a T5 model \cite{JMLR:v21:20-074} in the answer-aware setting. Given the input $\mathcal{D}^a_n = \{C$, $H_n$, $a_n$, $r_n$\} with $C$, $H_n$, $a_n$, $r_n$ as the context, ground-truth conversational history, ground-truth answer, and round-truth rationale, respectively, we conduct three experiments in \Cref{tab:question-yes-no-generation-performance}: original input with \texttt{Yes/No} keyword (\emph{With \texttt{Y/N}}), original input without \texttt{Yes/No} keyword (\emph{W/o \texttt{Y/N}}), original input without \texttt{Yes/No} and with the ground-truth $control\_signal$ (\emph{W/o \texttt{Y/N} + $control\_signal$}). Note that we train the model with the whole context, and a maximum of three previous history turns, as discussed in \Cref{appendix:question-generation-performance}. We measure the performance of the answer-aware CQG model separately on two types of questions: boolean and span-based by ROUGE-L \cite{lin-2004-rouge} and BERTScore \cite{Zhang2020BERTScore:}.

\paragraph{Observations} 
\Cref{tab:question-yes-no-generation-performance} shows the experimental results. We derive two main observations. Firstly, without knowing the keyword \texttt{Yes/No} (\emph{W/o \texttt{Y/N}}) - \emph{this is the case in the answer-unaware setting}, the model performs worse. This decrease shows that the \texttt{Yes/No} keyword is indeed helpful in hinting the model towards generating the correct questions. Secondly, by inputting the ground-truth $control\_signal$ into the model (\emph{W/o \texttt{Y/N} + $control\_signal$}), the performance is improved by a large margin compared to (\emph{W/o \texttt{Y/N}}). We obtain three implications from the above improvement. Firstly, it consolidates our hypothesis that inputting the ground-truth $control\_signal$ is truly helpful. Secondly, by training with the $control\_signal$, the performance of the model is even higher than with \texttt{Y/N} in the span-based cases, which indicates that training the model with $control\_signal$ makes it more stable to generate the correct questions. Thirdly, the performance of (\emph{W/o \texttt{Y/N} + $control\_signal$}) is lower than (\emph{With \texttt{Y/N}}) in boolean cases. The reason is \texttt{<BOOLEAN>} only informs the model to generate a boolean question without informing to generate an \texttt{Yes} or \texttt{No} one.

\section{Conclusion} \label{sec:conclusion}

This paper presents SG-CQG, a two-stage framework for the CQG task in the answer-unaware setting. Firstly, the \emph{what-to-ask} module aims to select a sentence as the rationale by the proposed semantic graph and extract the answer span from it. The \emph{how-to-ask} module classifies the 
type 
of the question before generating and filtering it. Additionally, we propose a set of automatic evaluation criteria for answer-unaware CQG, especially a novel metric, \emph{Conv-Distinct}, to evaluate the generated conversation from a context. Extensive automatic evaluation and human evaluation show that our method achieves state-of-the-art performances in the answer-unaware setting on CoQA, with a significant improvement in the conversational alignment property compared to previous frameworks. In the future, we will focus on how to reason over our semantic graph to select the rationale, and further improve the performances of how-to-ask module.

\section*{Limitations}
A limitation of our work is that our Graph Traversal Algorithm (Section 3.1) is a heuristic and unlearned algorithm. This leads to a number of nodes after being selected by this algorithm are not suitable for the model to generate conversational questions, and are eventually filtered out by other modules. Future works can focus on more advanced techniques to guide the model to select the nodes such as Graph Neural Networks \cite{wu2020comprehensive}. Furthermore, our algorithm to select the relevant turns in the conversational history to generate the conversational questions is a heuristic of selecting a maximum of three previous turns. This heuristic may not be optimal for the model to gather necessary information from history to generate conversational questions in the next turns, as discussed by \citet{cohs-cqg}. 

\section*{Ethical Considerations}
In this paper, we present a two-stage CQG framework (SG-CQG), which was trained on CoQA \cite{reddy-etal-2019-coqa}, a published large-scale dataset for building Conversational Question Answering systems. Our framework is potentially helpful for building chatbot systems, which can serve different streams such as educational, medical, or commercial purposes. 

Through human evaluations, we observe that our proposed method does not generate any discriminatory, insulting responses (questions and answers). We validate the proposed method and baseline models on human evaluation which involves manual labor. We hire three annotators to score 125 generated questions in total. The hourly pay is set to S\$15, which is higher than the local statutory minimum wage. Therefore, we do not anticipate any major ethical concerns.

\section*{Acknowledgements}
This research has been supported by the Institute for Infocomm Research of A*STAR (CR-2021-001). We would like to thank anonymous reviewers from ARR for their valuable feedback which helped us to improve our paper. We also want to thank Dr. Richeng Duan (A*STAR) for his feedback in the initial stage of the project.

\bibliography{anthology,custom}

\begin{thebibliography}{43}
\expandafter\ifx\csname natexlab\endcsname\relax\def\natexlab#1{#1}\fi

\bibitem[{Adiwardana et~al.(2020)Adiwardana, Luong, So, Hall, Fiedel,
  Thoppilan, Yang, Kulshreshtha, Nemade, Lu et~al.}]{adiwardana2020towards}
Daniel Adiwardana, Minh-Thang Luong, David~R So, Jamie Hall, Noah Fiedel, Romal
  Thoppilan, Zi~Yang, Apoorv Kulshreshtha, Gaurav Nemade, Yifeng Lu, et~al.
  2020.
\newblock Towards a human-like open-domain chatbot.
\newblock \emph{arXiv preprint arXiv:2001.09977}.

\bibitem[{Alberti et~al.(2019)Alberti, Andor, Pitler, Devlin, and
  Collins}]{alberti-etal-2019-synthetic}
Chris Alberti, Daniel Andor, Emily Pitler, Jacob Devlin, and Michael Collins.
  2019.
\newblock \href {https://doi.org/10.18653/v1/P19-1620} {Synthetic {QA} corpora
  generation with roundtrip consistency}.
\newblock In \emph{Proceedings of the 57th Annual Meeting of the Association
  for Computational Linguistics}, pages 6168--6173, Florence, Italy.
  Association for Computational Linguistics.

\bibitem[{Allen et~al.(2007)Allen, Chambers, Ferguson, Galescu, Jung, Swift,
  and Taysom}]{James07}
James Allen, Nathanael Chambers, George Ferguson, Lucian Galescu, Hyuckchul
  Jung, Mary Swift, and William Taysom. 2007.
\newblock Plow: A collaborative task learning agent.
\newblock In \emph{Proceedings of the 22nd National Conference on Artificial
  Intelligence - Volume 2}, AAAI'07, page 1514–1519. AAAI Press.

\bibitem[{Dagan et~al.(2006)Dagan, Glickman, and Magnini}]{10.1007/11736790_9}
Ido Dagan, Oren Glickman, and Bernardo Magnini. 2006.
\newblock The pascal recognising textual entailment challenge.
\newblock In \emph{Machine Learning Challenges. Evaluating Predictive
  Uncertainty, Visual Object Classification, and Recognising Tectual
  Entailment}, pages 177--190, Berlin, Heidelberg. Springer Berlin Heidelberg.

\bibitem[{Do et~al.(2022)Do, Zou, Pan, Chen, Joty, and Aw}]{cohs-cqg}
Xuan~Long Do, Bowei Zou, Liangming Pan, Nancy~F. Chen, Shafiq Joty, and Ai~Ti
  Aw. 2022.
\newblock \href {https://aclanthology.org/2022.coling-1.48} {{C}o{HS}-{CQG}:
  Context and history selection for conversational question generation}.
\newblock In \emph{Proceedings of the 29th International Conference on
  Computational Linguistics}, pages 580--591, Gyeongju, Republic of Korea.
  International Committee on Computational Linguistics.

\bibitem[{Dziri et~al.(2019)Dziri, Kamalloo, Mathewson, and
  Zaiane}]{dziri-etal-2019-evaluating-coherence}
Nouha Dziri, Ehsan Kamalloo, Kory Mathewson, and Osmar Zaiane. 2019.
\newblock \href {https://aclanthology.org/W19-3646} {Evaluating coherence in
  dialogue systems using entailment}.
\newblock In \emph{Proceedings of the 2019 Workshop on Widening NLP}, pages
  146--148, Florence, Italy. Association for Computational Linguistics.

\bibitem[{Gao et~al.(2019)Gao, Li, King, and
  Lyu}]{gao-etal-2019-interconnected}
Yifan Gao, Piji Li, Irwin King, and Michael~R. Lyu. 2019.
\newblock \href {https://doi.org/10.18653/v1/P19-1480} {Interconnected question
  generation with coreference alignment and conversation flow modeling}.
\newblock In \emph{Proceedings of the 57th Annual Meeting of the Association
  for Computational Linguistics}, pages 4853--4862, Florence, Italy.
  Association for Computational Linguistics.

\bibitem[{Ghazarian et~al.(2019)Ghazarian, Wei, Galstyan, and
  Peng}]{ghazarian-etal-2019-better}
Sarik Ghazarian, Johnny Wei, Aram Galstyan, and Nanyun Peng. 2019.
\newblock \href {https://doi.org/10.18653/v1/W19-2310} {Better automatic
  evaluation of open-domain dialogue systems with contextualized embeddings}.
\newblock In \emph{Proceedings of the Workshop on Methods for Optimizing and
  Evaluating Neural Language Generation}, pages 82--89, Minneapolis, Minnesota.
  Association for Computational Linguistics.

\bibitem[{Gu et~al.(2021)Gu, Mirshekari, Yu, and Sisto}]{gu-etal-2021-chaincqg}
Jing Gu, Mostafa Mirshekari, Zhou Yu, and Aaron Sisto. 2021.
\newblock \href {https://doi.org/10.18653/v1/2021.eacl-main.177} {{C}hain{CQG}:
  Flow-aware conversational question generation}.
\newblock In \emph{Proceedings of the 16th Conference of the European Chapter
  of the Association for Computational Linguistics: Main Volume}, pages
  2061--2070, Online. Association for Computational Linguistics.

\bibitem[{Holtzman et~al.(2020)Holtzman, Buys, Du, Forbes, and
  Choi}]{Holtzman2020The}
Ari Holtzman, Jan Buys, Li~Du, Maxwell Forbes, and Yejin Choi. 2020.
\newblock \href {https://openreview.net/forum?id=rygGQyrFvH} {The curious case
  of neural text degeneration}.
\newblock In \emph{International Conference on Learning Representations}.

\bibitem[{Krippendorff(2011)}]{krippendorff2011computing}
Klaus Krippendorff. 2011.
\newblock Computing krippendorff's alpha-reliability.
\newblock \emph{Computing}, 1.

\bibitem[{Lewis et~al.(2020)Lewis, Liu, Goyal, Ghazvininejad, Mohamed, Levy,
  Stoyanov, and Zettlemoyer}]{lewis-etal-2020-bart}
Mike Lewis, Yinhan Liu, Naman Goyal, Marjan Ghazvininejad, Abdelrahman Mohamed,
  Omer Levy, Veselin Stoyanov, and Luke Zettlemoyer. 2020.
\newblock \href {https://doi.org/10.18653/v1/2020.acl-main.703} {{BART}:
  Denoising sequence-to-sequence pre-training for natural language generation,
  translation, and comprehension}.
\newblock In \emph{Proceedings of the 58th Annual Meeting of the Association
  for Computational Linguistics}, pages 7871--7880, Online. Association for
  Computational Linguistics.

\bibitem[{Li et~al.(2016{\natexlab{a}})Li, Galley, Brockett, Gao, and
  Dolan}]{li-etal-2016-diversity}
Jiwei Li, Michel Galley, Chris Brockett, Jianfeng Gao, and Bill Dolan.
  2016{\natexlab{a}}.
\newblock \href {https://doi.org/10.18653/v1/N16-1014} {A diversity-promoting
  objective function for neural conversation models}.
\newblock In \emph{Proceedings of the 2016 Conference of the North {A}merican
  Chapter of the Association for Computational Linguistics: Human Language
  Technologies}, pages 110--119, San Diego, California. Association for
  Computational Linguistics.

\bibitem[{Li et~al.(2016{\natexlab{b}})Li, Miller, Chopra, Ranzato, and
  Weston}]{li2016learning}
Jiwei Li, Alexander~H Miller, Sumit Chopra, Marc'Aurelio Ranzato, and Jason
  Weston. 2016{\natexlab{b}}.
\newblock Learning through dialogue interactions by asking questions.
\newblock \emph{International Conference on Learning Representations 2017}.

\bibitem[{Lin(2004)}]{lin-2004-rouge}
Chin-Yew Lin. 2004.
\newblock \href {https://aclanthology.org/W04-1013} {{ROUGE}: A package for
  automatic evaluation of summaries}.
\newblock In \emph{Text Summarization Branches Out}, pages 74--81, Barcelona,
  Spain. Association for Computational Linguistics.

\bibitem[{Liu et~al.(2019{\natexlab{a}})Liu, Ott, Goyal, Du, Joshi, Chen, Levy,
  Lewis, Zettlemoyer, and Stoyanov}]{roberta-paper}
Yinhan Liu, Myle Ott, Naman Goyal, Jingfei Du, Mandar Joshi, Danqi Chen, Omer
  Levy, Mike Lewis, Luke Zettlemoyer, and Veselin Stoyanov. 2019{\natexlab{a}}.
\newblock \href {http://arxiv.org/abs/1907.11692} {Roberta: {A} robustly
  optimized {BERT} pretraining approach}.
\newblock \emph{CoRR}, abs/1907.11692.

\bibitem[{Liu et~al.(2019{\natexlab{b}})Liu, Niu, Wu, and
  Wang}]{liu-etal-2019-knowledge}
Zhibin Liu, Zheng-Yu Niu, Hua Wu, and Haifeng Wang. 2019{\natexlab{b}}.
\newblock \href {https://doi.org/10.18653/v1/D19-1187} {Knowledge aware
  conversation generation with explainable reasoning over augmented graphs}.
\newblock In \emph{Proceedings of the 2019 Conference on Empirical Methods in
  Natural Language Processing and the 9th International Joint Conference on
  Natural Language Processing (EMNLP-IJCNLP)}, pages 1782--1792, Hong Kong,
  China. Association for Computational Linguistics.

\bibitem[{Loshchilov and Hutter(2019)}]{loshchilov2018decoupled}
Ilya Loshchilov and Frank Hutter. 2019.
\newblock \href {https://openreview.net/forum?id=Bkg6RiCqY7} {Decoupled weight
  decay regularization}.
\newblock In \emph{International Conference on Learning Representations}.

\bibitem[{Lu and Lu(2021)}]{lu-lu-2021-survey}
Chao-Yi Lu and Sin-En Lu. 2021.
\newblock \href {https://aclanthology.org/2021.rocling-1.21} {A survey of
  approaches to automatic question generation:from 2019 to early 2021}.
\newblock In \emph{Proceedings of the 33rd Conference on Computational
  Linguistics and Speech Processing (ROCLING 2021)}, pages 151--162, Taoyuan,
  Taiwan. The Association for Computational Linguistics and Chinese Language
  Processing (ACLCLP).

\bibitem[{Moghe et~al.(2018)Moghe, Arora, Banerjee, and
  Khapra}]{moghe-etal-2018-towards}
Nikita Moghe, Siddhartha Arora, Suman Banerjee, and Mitesh~M. Khapra. 2018.
\newblock \href {https://doi.org/10.18653/v1/D18-1255} {Towards exploiting
  background knowledge for building conversation systems}.
\newblock In \emph{Proceedings of the 2018 Conference on Empirical Methods in
  Natural Language Processing}, pages 2322--2332, Brussels, Belgium.
  Association for Computational Linguistics.

\bibitem[{Mostafazadeh et~al.(2016)Mostafazadeh, Chambers, He, Parikh, Batra,
  Vanderwende, Kohli, and Allen}]{mostafazadeh-etal-2016-corpus}
Nasrin Mostafazadeh, Nathanael Chambers, Xiaodong He, Devi Parikh, Dhruv Batra,
  Lucy Vanderwende, Pushmeet Kohli, and James Allen. 2016.
\newblock \href {https://doi.org/10.18653/v1/N16-1098} {A corpus and cloze
  evaluation for deeper understanding of commonsense stories}.
\newblock In \emph{Proceedings of the 2016 Conference of the North {A}merican
  Chapter of the Association for Computational Linguistics: Human Language
  Technologies}, pages 839--849, San Diego, California. Association for
  Computational Linguistics.

\bibitem[{Nakanishi et~al.(2019)Nakanishi, Kobayashi, and
  Hayashi}]{nakanishi-etal-2019-towards}
Mao Nakanishi, Tetsunori Kobayashi, and Yoshihiko Hayashi. 2019.
\newblock \href {https://doi.org/10.18653/v1/D19-5809} {Towards answer-unaware
  conversational question generation}.
\newblock In \emph{Proceedings of the 2nd Workshop on Machine Reading for
  Question Answering}, pages 63--71, Hong Kong, China. Association for
  Computational Linguistics.

\bibitem[{Ott et~al.(2019)Ott, Edunov, Baevski, Fan, Gross, Ng, Grangier, and
  Auli}]{ott2019fairseq}
Myle Ott, Sergey Edunov, Alexei Baevski, Angela Fan, Sam Gross, Nathan Ng,
  David Grangier, and Michael Auli. 2019.
\newblock fairseq: A fast, extensible toolkit for sequence modeling.
\newblock In \emph{Proceedings of NAACL-HLT 2019: Demonstrations}.

\bibitem[{Pan et~al.(2019{\natexlab{a}})Pan, Li, Yao, Cai, and
  Sun}]{pan-etal-2019-reinforced}
Boyuan Pan, Hao Li, Ziyu Yao, Deng Cai, and Huan Sun. 2019{\natexlab{a}}.
\newblock \href {https://doi.org/10.18653/v1/P19-1203} {Reinforced dynamic
  reasoning for conversational question generation}.
\newblock In \emph{Proceedings of the 57th Annual Meeting of the Association
  for Computational Linguistics}, pages 2114--2124, Florence, Italy.
  Association for Computational Linguistics.

\bibitem[{Pan et~al.(2019{\natexlab{b}})Pan, Lei, Chua, and
  Kan}]{https://doi.org/10.48550/arxiv.1905.08949}
Liangming Pan, Wenqiang Lei, Tat-Seng Chua, and Min-Yen Kan.
  2019{\natexlab{b}}.
\newblock \href {https://doi.org/10.48550/ARXIV.1905.08949} {Recent advances in
  neural question generation}.

\bibitem[{Pang et~al.(2020)Pang, Nijkamp, Han, Zhou, Liu, and
  Tu}]{pang-etal-2020-towards}
Bo~Pang, Erik Nijkamp, Wenjuan Han, Linqi Zhou, Yixian Liu, and Kewei Tu. 2020.
\newblock \href {https://doi.org/10.18653/v1/2020.acl-main.333} {Towards
  holistic and automatic evaluation of open-domain dialogue generation}.
\newblock In \emph{Proceedings of the 58th Annual Meeting of the Association
  for Computational Linguistics}, pages 3619--3629, Online. Association for
  Computational Linguistics.

\bibitem[{Puri et~al.(2020)Puri, Spring, Shoeybi, Patwary, and
  Catanzaro}]{puri-etal-2020-training}
Raul Puri, Ryan Spring, Mohammad Shoeybi, Mostofa Patwary, and Bryan Catanzaro.
  2020.
\newblock \href {https://doi.org/10.18653/v1/2020.emnlp-main.468} {Training
  question answering models from synthetic data}.
\newblock In \emph{Proceedings of the 2020 Conference on Empirical Methods in
  Natural Language Processing (EMNLP)}, pages 5811--5826, Online. Association
  for Computational Linguistics.

\bibitem[{Qi et~al.(2020)Qi, Zhang, and Manning}]{qi-etal-2020-stay}
Peng Qi, Yuhao Zhang, and Christopher~D. Manning. 2020.
\newblock \href {https://doi.org/10.18653/v1/2020.findings-emnlp.3} {Stay
  hungry, stay focused: Generating informative and specific questions in
  information-seeking conversations}.
\newblock In \emph{Findings of the Association for Computational Linguistics:
  EMNLP 2020}, pages 25--40, Online. Association for Computational Linguistics.

\bibitem[{Radford et~al.(2019)Radford, Wu, Child, Luan, Amodei, and
  Sutskever}]{radford2019language}
Alec Radford, Jeff Wu, Rewon Child, David Luan, Dario Amodei, and Ilya
  Sutskever. 2019.
\newblock Language models are unsupervised multitask learners.

\bibitem[{Raffel et~al.(2020)Raffel, Shazeer, Roberts, Lee, Narang, Matena,
  Zhou, Li, and Liu}]{JMLR:v21:20-074}
Colin Raffel, Noam Shazeer, Adam Roberts, Katherine Lee, Sharan Narang, Michael
  Matena, Yanqi Zhou, Wei Li, and Peter~J. Liu. 2020.
\newblock \href {http://jmlr.org/papers/v21/20-074.html} {Exploring the limits
  of transfer learning with a unified text-to-text transformer}.
\newblock \emph{Journal of Machine Learning Research}, 21(140):1--67.

\bibitem[{Rajpurkar et~al.(2016)Rajpurkar, Zhang, Lopyrev, and
  Liang}]{rajpurkar-etal-2016-squad}
Pranav Rajpurkar, Jian Zhang, Konstantin Lopyrev, and Percy Liang. 2016.
\newblock \href {https://doi.org/10.18653/v1/D16-1264} {{SQ}u{AD}: 100,000+
  questions for machine comprehension of text}.
\newblock In \emph{Proceedings of the 2016 Conference on Empirical Methods in
  Natural Language Processing}, pages 2383--2392, Austin, Texas. Association
  for Computational Linguistics.

\bibitem[{Reddy et~al.(2019)Reddy, Chen, and Manning}]{reddy-etal-2019-coqa}
Siva Reddy, Danqi Chen, and Christopher~D. Manning. 2019.
\newblock \href {https://doi.org/10.1162/tacl_a_00266} {{C}o{QA}: A
  conversational question answering challenge}.
\newblock \emph{Transactions of the Association for Computational Linguistics},
  7:249--266.

\bibitem[{Shakeri et~al.(2020)Shakeri, Nogueira~dos Santos, Zhu, Ng, Nan, Wang,
  Nallapati, and Xiang}]{shakeri-etal-2020-end}
Siamak Shakeri, Cicero Nogueira~dos Santos, Henghui Zhu, Patrick Ng, Feng Nan,
  Zhiguo Wang, Ramesh Nallapati, and Bing Xiang. 2020.
\newblock \href {https://doi.org/10.18653/v1/2020.emnlp-main.439} {End-to-end
  synthetic data generation for domain adaptation of question answering
  systems}.
\newblock In \emph{Proceedings of the 2020 Conference on Empirical Methods in
  Natural Language Processing (EMNLP)}, pages 5445--5460, Online. Association
  for Computational Linguistics.

\bibitem[{Shen et~al.(2021)Shen, Meng, Zhang, Feng, and
  Zhou}]{shen-etal-2021-gtm}
Lei Shen, Fandong Meng, Jinchao Zhang, Yang Feng, and Jie Zhou. 2021.
\newblock \href {https://doi.org/10.18653/v1/2021.acl-long.271} {{GTM}: A
  generative triple-wise model for conversational question generation}.
\newblock In \emph{Proceedings of the 59th Annual Meeting of the Association
  for Computational Linguistics and the 11th International Joint Conference on
  Natural Language Processing (Volume 1: Long Papers)}, pages 3495--3506,
  Online. Association for Computational Linguistics.

\bibitem[{Shi and Lin(2019)}]{https://doi.org/10.48550/arxiv.1904.05255}
Peng Shi and Jimmy Lin. 2019.
\newblock \href {https://doi.org/10.48550/ARXIV.1904.05255} {Simple bert models
  for relation extraction and semantic role labeling}.

\bibitem[{Wang et~al.(2018)Wang, Liu, Huang, and
  Nie}]{wang-etal-2018-learning-ask}
Yansen Wang, Chenyi Liu, Minlie Huang, and Liqiang Nie. 2018.
\newblock \href {https://doi.org/10.18653/v1/P18-1204} {Learning to ask
  questions in open-domain conversational systems with typed decoders}.
\newblock In \emph{Proceedings of the 56th Annual Meeting of the Association
  for Computational Linguistics (Volume 1: Long Papers)}, pages 2193--2203,
  Melbourne, Australia. Association for Computational Linguistics.

\bibitem[{Wolf et~al.(2020)Wolf, Debut, Sanh, Chaumond, Delangue, Moi, Cistac,
  Rault, Louf, Funtowicz, Davison, Shleifer, von Platen, Ma, Jernite, Plu, Xu,
  Le~Scao, Gugger, Drame, Lhoest, and Rush}]{wolf-etal-2020-transformers}
Thomas Wolf, Lysandre Debut, Victor Sanh, Julien Chaumond, Clement Delangue,
  Anthony Moi, Pierric Cistac, Tim Rault, Remi Louf, Morgan Funtowicz, Joe
  Davison, Sam Shleifer, Patrick von Platen, Clara Ma, Yacine Jernite, Julien
  Plu, Canwen Xu, Teven Le~Scao, Sylvain Gugger, Mariama Drame, Quentin Lhoest,
  and Alexander Rush. 2020.
\newblock \href {https://doi.org/10.18653/v1/2020.emnlp-demos.6} {Transformers:
  State-of-the-art natural language processing}.
\newblock In \emph{Proceedings of the 2020 Conference on Empirical Methods in
  Natural Language Processing: System Demonstrations}, pages 38--45, Online.
  Association for Computational Linguistics.

\bibitem[{Wu et~al.(2020)Wu, Pan, Chen, Long, Zhang, and
  Philip}]{wu2020comprehensive}
Zonghan Wu, Shirui Pan, Fengwen Chen, Guodong Long, Chengqi Zhang, and S~Yu
  Philip. 2020.
\newblock A comprehensive survey on graph neural networks.
\newblock \emph{IEEE transactions on neural networks and learning systems},
  32(1):4--24.

\bibitem[{Xu et~al.(2020)Xu, Lei, Wang, Niu, Wu, and Che}]{ijcai2020p545}
Jun Xu, Zeyang Lei, Haifeng Wang, Zheng-Yu Niu, Hua Wu, and Wanxiang Che. 2020.
\newblock \href {https://doi.org/10.24963/ijcai.2020/545} {Enhancing dialog
  coherence with event graph grounded content planning}.
\newblock In \emph{Proceedings of the Twenty-Ninth International Joint
  Conference on Artificial Intelligence, {IJCAI-20}}, pages 3941--3947.
  International Joint Conferences on Artificial Intelligence Organization.
\newblock Main track.

\bibitem[{Xu et~al.(2021)Xu, Lei, Wang, Niu, Wu, and
  Che}]{xu-etal-2021-discovering}
Jun Xu, Zeyang Lei, Haifeng Wang, Zheng-Yu Niu, Hua Wu, and Wanxiang Che. 2021.
\newblock \href {https://doi.org/10.18653/v1/2021.acl-long.136} {Discovering
  dialog structure graph for coherent dialog generation}.
\newblock In \emph{Proceedings of the 59th Annual Meeting of the Association
  for Computational Linguistics and the 11th International Joint Conference on
  Natural Language Processing (Volume 1: Long Papers)}, pages 1726--1739,
  Online. Association for Computational Linguistics.

\bibitem[{Yeh et~al.(2021)Yeh, Eskenazi, and
  Mehri}]{yeh-etal-2021-comprehensive}
Yi-Ting Yeh, Maxine Eskenazi, and Shikib Mehri. 2021.
\newblock \href {https://doi.org/10.18653/v1/2021.eancs-1.3} {A comprehensive
  assessment of dialog evaluation metrics}.
\newblock In \emph{The First Workshop on Evaluations and Assessments of Neural
  Conversation Systems}, pages 15--33, Online. Association for Computational
  Linguistics.

\bibitem[{Yue et~al.(2022)Yue, Yao, and Sun}]{yue-etal-2022-synthetic}
Xiang Yue, Ziyu Yao, and Huan Sun. 2022.
\newblock \href {https://doi.org/10.18653/v1/2022.acl-long.95} {Synthetic
  question value estimation for domain adaptation of question answering}.
\newblock In \emph{Proceedings of the 60th Annual Meeting of the Association
  for Computational Linguistics (Volume 1: Long Papers)}, pages 1340--1351,
  Dublin, Ireland. Association for Computational Linguistics.

\bibitem[{Zhang et~al.(2020)Zhang, Kishore, Wu, Weinberger, and
  Artzi}]{Zhang2020BERTScore:}
Tianyi Zhang, Varsha Kishore, Felix Wu, Kilian~Q. Weinberger, and Yoav Artzi.
  2020.
\newblock \href {https://openreview.net/forum?id=SkeHuCVFDr} {Bertscore:
  Evaluating text generation with bert}.
\newblock In \emph{International Conference on Learning Representations}.

\end{thebibliography}
\bibliographystyle{acl_natbib}

\clearpage

\appendix
\section{Appendix} \label{sec:appendix}

\subsection{Extended Related Work} \label{appendix:extended-related-work}
Our work is related to two more lines of prior work.
\subsubsection{Synthetic Question-Answering (QA) Generation}
Synthetic QA generation based on pretrained language models (LM) has been studied and demonstrated the helpfulness in improving the downstream Reading Comprehension (RC) task \cite{alberti-etal-2019-synthetic, puri-etal-2020-training, shakeri-etal-2020-end}. \citet{alberti-etal-2019-synthetic} proposed a novel method to generate synthetic data by combining models of question generation and answer extraction, and by filtering the results to ensure roundtrip consistency. However, this work differs from ours since it only considered the task of single-turn QA generation and focused only on extractive QA generation while we focus on multi-turn QA generation and have both span-based and boolean questions. Regarding the filtering technique, this work and \citet{puri-etal-2020-training} used \emph{round-trip filtering} method, which is similar to \citet{cohs-cqg} and is a relaxed version of our filtering module.  \citet{shakeri-etal-2020-end} later introduced an end-to-end framework to generate QA data. This work used \emph{LM filtering} method, which is similar to \emph{sample-and-reranking} \cite{Holtzman2020The} and ours. In our case (as discussed in \emph{(1) Wrong answer} error in \Cref{how-to-ask}), to filter QA pairs, we also sample multiple answers from a QA model and select the answers with the highest frequency and confidence score by the model. If the highest frequency one is different from the highest confidence one, we filter our the question.

\subsubsection{Dialog Generation Evaluation}
Dialog evaluation metrics have been studied extensively \cite{yeh-etal-2021-comprehensive}. However, it is worth noting that this task is different from ours, since we prefer evaluating the questions in QA conversations only. In addition, when conducting experiments with reference-free dialog generation metrics like BERT-RUBER \cite{ghazarian-etal-2019-better} and HolisticEval \cite{pang-etal-2020-towards}, we observe that these metrics are not suitable for evaluating QA pairs since the questions and answers in QA conversations are normally shorter without many referential details among turns compared to dialog responses.

Previous works \cite{alberti-etal-2019-synthetic, puri-etal-2020-training, shakeri-etal-2020-end} usually evaluated the generated QA data by training the RC systems with it and examining whether the synthetic data improves the RC systems without actually examining the synthetic data. Recent work \cite{cohs-cqg} evaluated the QA pairs manually. In addition, \cite{yue-etal-2022-synthetic} proposed \emph{question value estimator}, a novel module to estimate the usefulness of synthetic questions to improve the target-domain QA performance. However, this is not directly relevant to ours since even though the metric can evaluate the usefulness of the generated questions, it does not offer the actual properties of the generated questions. To the best of our knowledge, our work is the first one that proposes a set of criteria to evaluate the question-answer pairs in QA conversations. The performance of models evaluated by our proposed automatic evaluation metrics (\Cref{tab:main-evaluation}) is positively correlated with human evaluation (\Cref{tab:human-evaluation}) where we observe that improvements on our metrics are also improvements on human evaluation metrics.

\subsection{Graph Traversal Algorithm} \label{appendix:graph-traversal-algorithm}

We present the pseudocode of our \emph{Graph Traversal Algorithm}, which is described in \Cref{what-to-ask}. 

\begin{algorithm}
\DontPrintSemicolon
\caption{Graph Traversal Algorithm} \label{alg:graph-traversal}
  \KwInput{$\mathcal{G} = \{\mathcal{V}, \mathcal{E}\}$, $H = \{(q_1, a_1), ..., (q_{n-1}, a_{n-1})\}$/$\emptyset$.}
  \KwOutput{Index of the sentence as $r_n$}
  \KwInitialize{$I$: nodes in rationales of $H$, \; \, $q$: queue of nodes to visit, \;
  \, Add nodes in $I$ to $q$ in the index order.
  }
   \While{q is not empty}
   {
   		$cur$ = $q$[0] \;
   		del $q$[0] \\
   		\If{cur is visited twice}
   		{
   		    continue
   		}
   		$r_n$ = retrieve sentence contains $q$[0] \\
   		$A_n$ = answer spans set extracted from $r_n$ \\
   		\eIf{successfully generate $q_n$ from $r_n$ and any $a_n \in A$}
        {
            Add unvisited neighbors of $cur$ to the beginning of $q$  
        }{
        	Add unvisited neighbors of $cur$ to the end of $q$   
        }
   }  
\end{algorithm}
\vspace{-3mm}

\subsection{Question Generation} \label{appendix:question-generation-performance}
Given the input $\mathcal{D}^a_n = \{C$, $H_n$, $a_n$, $r_n$, $control\_signal\}$ in which $C$, $H_n$, $a_n$, $r_n$, $control\_signal$ are the context, conversational history, expected answer, rationale, and the control signal respectively, we fine-tune a T5$_{base}$ model \cite{JMLR:v21:20-074} as our question generation model. \citet{cohs-cqg} showed that by training the T5 model with the whole context and the shortened conversational history, the performance of the model is improved. We replicate this experiment by reporting the performance of the T5 model with a different number of the previous history turns in \Cref{tab:ablation-previous-turn}. We derive the same observation as \citet{cohs-cqg}, which is the model performs the best with a maximum of two or three conversational previous turns. As such, we opt for selecting at most 3 previous turns to train our QG model.

\begin{table}[!t]
\centering
\resizebox{0.45\textwidth}{!}{%
\begin{tabular}{cccc}
\hline
\emph{\#Pre. turns} & ROUGE-L & BLEU-4 & BERTScore \\
\hline
1 & 48.64 & 17.93 & 93.42 \\
2 & 48.77 & \textbf{18.27} & 93.43 \\
3 & \textbf{48.84} & 18.18 & \textbf{93.46} \\
4 & 48.27 & 18.16 & 93.38 \\
\hline
Full history & 45.93 & 17.11 & 93.09 \\
\hline
\end{tabular}
}
\vspace{3mm}
\caption{\small{Performance of the T5 model, training with different fixed number of previous turns on our validation set.}}
\label{tab:ablation-previous-turn}
\end{table}

\subsection{Adding Extra Edges Algorithm} \label{appendix:component-merging-algorithm}
We provide the pseudocode for the adding-\emph{Extra}-edges algorithm in \Cref{algo:merging-connected-component-algorithm}.

\begin{algorithm}
\DontPrintSemicolon
\caption{Adding Extra Edges} \label{algo:merging-connected-component-algorithm}  
     \KwInput{$\mathcal{G} = \{(u, v)$\} for u, v are nodes in directed graph that belong to the same sentence. For different sentences, only consider the starting node and the ending node.\;  
     }
    \KwOutput{The set of newly added edges} 
    \KwInitialize{A disjoint set union (DSU) for checking whether 2 sentences are in the same component.} 

     \ $addedEdges$ = [] \;
     \ $pairs$ = all pairs of 2 sentences \;
     \ sort($pairs$)  \tcp{for prioritizing those pairs with the minimum index difference} 
    \For {$pair$ in $pairs$}{
    $p_1$, $p_2$ = $pair$[0], $pair$[1] \;
    $sameComponent$ = check the connectivity of $p_1$, $p_2$ by DSU\; 
    \If{not $sameComponent$}
    {
        merge sentences $p_1$ and $p_2$ into the same component by DSU \;
        add new edge between the ending node of sentence $p_1$ with starting node of sentence $p_2$ to $addedEdges$ \;
    }
    }
return $addedEdges$
\end{algorithm}

\subsection{Details of Question Type Classifier} \label{appendix:question-type-classifier-input}
In this section, we detail our setting to train and validate the proposed \emph{Question Type Classifier}. We conduct our experiments on train set, our test set (i.e. CoQA validation set) and our validation set of CoQA \cite{reddy-etal-2019-coqa}. For each conversation, we automatically label its questions according to their answers. In particular, a question is labeled as boolean if its answer begins with \texttt{Yes/No/yes/no/YES/NO}, and span-based otherwise. Given the input $\mathcal{D}^a_n = \{C$, $H_n$, $a_n$, $r_n$\} with $C$, $H_n$, $a_n$, $r_n$ are the context, ground-truth conversational history,  ground-truth answer,  round-truth rationale respectively, we construct the input to the classifier as followed. If $a_n \in \{\texttt{Yes, No, yes, no, YES, NO}\}$, the input to the classifier is \texttt{Answer}: \textit{$r_n$} \textit{$r_n$} \texttt{Context:} \textit{$C$} \texttt{[SEP]} $H_{sub}$, else, the input is \texttt{Answer}: \textit{$a_n$} \textit{$r_n$} \textit{$r_n$} \texttt{Context:} \textit{$C$} \texttt{[SEP]} where $H_{sub}$ is the shortened $H_n$, in which we keep at most three previous turns, and the output is \texttt{0/1} indicating whether the ground-truth question is boolean/span-based. Our classifier achieves an accuracy of 95.6\%.

\subsection{Details of CQA Model} \label{appendix:details-cqa-model}
We fine-tuned a T5 \cite{JMLR:v21:20-074} as our Conversational Question Answering (CQA) model on CoQA \cite{reddy-etal-2019-coqa}. The input to the model follows the format: \texttt{Question: Q [SEP] Context: C [SEP] H\_sub} in which \texttt{Q, C} are the question and the context respectively, and \texttt{H\_sub} is the shortened conversational history with a maximum of 3 previous turns. Our CQA model achieves $63.65\%$ Exact Match (EM) and $74.08\%$ F1, as we presented in \Cref{tab:answer-evaluation}.

\subsection{Evaluation Metrics Discussion} \label{appendix:evaluation-metric-calculation}

One of our core contributions is the set of criteria to evaluate question-answer conversations. In this section, we detail our intuitions as well as computations of the metrics.

\subsubsection{Distinct-N \cite{li-etal-2016-diversity}}
Distinct-N \cite{li-etal-2016-diversity} is a N-gram metric to measure the diversity of a sentence. In our experiments, we calculate Distinct-1 score and Distinct-2 score provided by \citet{li-etal-2016-diversity}\footnote{https://github.com/neural-dialogue-metrics/Distinct-N}.

\subsubsection{Context Coverage and Conv-Distinct}
As we discussed in \Cref{sec:system-experiment}, one critical shortcoming when directly applying Distinct-N to evaluate the QA conversations is that the conversations with very few turns tend to attain very high Distinct-N scores. To address this challenge, we introduce Context Coverage (CC) and Conv-Distinct.   

\paragraph{Context Coverage (CC)} is measured as the percentage of sentences that are rationales. For example, given a context of 6 sentences, among them, 5 sentences are selected as rationales for a generated conversation. Then the CC score of this generated conversation is 5/6 = 0.84.

To compute CC Scores for E2E models, we classify a sentence as a rationale if there is at least one question-answer pair generated from that sentence. As a result, the model of \citet{cohs-cqg} and our \emph{SG-CQG} can output which sentence is a rationale, and it is straightforward to compute the CC scores. However, the end-to-end outputs of BART \cite{lewis-etal-2020-bart} and T5 \cite{JMLR:v21:20-074} are the question-answer pairs only, it is needed to find which sentence is a rationale of each pair. To do so, we adopt a simple heuristic. For each generated question-answer pair, we classify a sentence as its rationale if that sentence has the \emph{longest common substring} with the concatenation of its question and answer among all the sentences in the context. By that, we get the set of sentences that are rationales.

\paragraph{Conv-Distinct} is defined as the multiplication of the Distinct score of the
generated conversation with its CC score. For example, in the above generated conversation, the Distinct-1 score is 60.50. So its Conv-Distinct-1 score is 60.50 * 0.84 = 50.42.

It is worth noting that the diversity in token level is a common property of the dialog which has been discussed in many previous works \cite{qi-etal-2020-stay, pang-etal-2020-towards, adiwardana2020towards}.

\subsubsection{BERTScore$-$entailment $1,2,3$}
BERTScore$-$entailment is an upgraded version of Dialog$-$entailment metric \cite{dziri-etal-2019-evaluating-coherence}, which measures the topic coherence property by deep contextual representation. We follow \citet{dziri-etal-2019-evaluating-coherence} to characterize the consistency of dialogue systems as a natural language inference (NLI) problem \cite{10.1007/11736790_9}. This property is important for questions and answers in the QA conversation, because the questions should focus on the topic of previous turns, and the answers should focus on their questions. In our experiments, we compute BERTScore$-$entailment with $1, 2$, and all previous turn(s). The BERTScore calculation is adopted from its authors\footnote{https://github.com/Tiiiger/bert\_score}.

\subsubsection{BERTScore \cite{Zhang2020BERTScore:}}
We observe that Distinct-N, Conv-Distinct-N and BERTScore-entailment are only to measure the quality of the QA pairs. None of them measures the relationship between QA pairs and the given context. As such, we propose to use BERTScore \cite{Zhang2020BERTScore:} to measure the similarity of the generated conversation and the given context. It is worth noting that this metric only serves for measuring the similarity between the generated conversation and context only. A generated conversation with a very high similarity score with the given context does not reflect that it is a very good conversation, as in the case of BART \cite{lewis-etal-2020-bart} in \Cref{tab:main-evaluation}. We provide this metric to give audiences "a sense" of how the generated conversation is explicitly relevant to the given context.

\subsubsection{EM \& F1 Answerability Measurements} 
The Exact Match (EM) and F1 measurements in \Cref{sec:experimental-results} are to evaluate the answerability and the correctness of our generated questions and answers respectively (i.e. the quality of the generated conversational answers). Since from a context, multiple conversations can be generated, we argue that one critical aspect of a good conversation is the quality of the generated conversational answers, i.e. the conversational questions must be answerable by the given context, and their answers must be exactly the generated conversational answers.

\subsubsection{Jumping Score (JS)} \label{appendix:compute-js-score}
To further understand the characteristics of each model in generating conversations, we measure its jumping score. We define this score as the percentage of turns in which the model jumps back to any previous content of their previous turn (i.e. traceback). For example, a generated conversation with the indexes of rationales [1,4,3,5,8,6] has the JS score is 2/5 = 0.4. It has 2 turns (over a maximum of 5 jumping back turns) in which the model jumps back, which are the $3-$rd turn and $6-$th turn. It is worth noting that the JS only shows one of the aspects of the result analysis. We could not say a system with the highest JS is better than others. JS only reflects a kind of flexibility for a what-to-ask module to some extent. 
We observe that our proposed SG-CQG achieves the highest JS score, which reflects that our proposed \emph{what-to-ask} module is the most flexible in terms of selecting the sentences in the context. 



\subsection{Statistical Significance of Results} \label{appendix:statistical-significance}
We compute the Student’s t-test to measure the significant difference between our model’s performance and the best baseline for each evaluation metric with the null hypothesis \texttt{H0: There is no significant difference}, and \texttt{H1: There is a significant difference}. We obtained the p-values as in \Cref{tab:main-evaluation}:

$\bullet$ Compared to T5: 4.32e-11 (BERT-entailment all), 5.20e-98, (BERT-entailment 1), 2.48e-34 (BERT-entailment 2).

$\bullet$ Compared to CoHS-CQG: 7.62e-188 (CC Score), 5.12e-119 (Conv-Distinct 1), 8.11e-173 (Conv-Distinct 2). The p-values, in this case, are too small because the improvements are intuitively significant.

We observe that all the p-values are less than .01, which indicates that our improvements on those metrics are significant.

\subsection{Human Evaluation Scoring System} \label{appendix:human-rating-system}

\begin{figure}[t!]
\centering

\includegraphics[width=7.5cm]{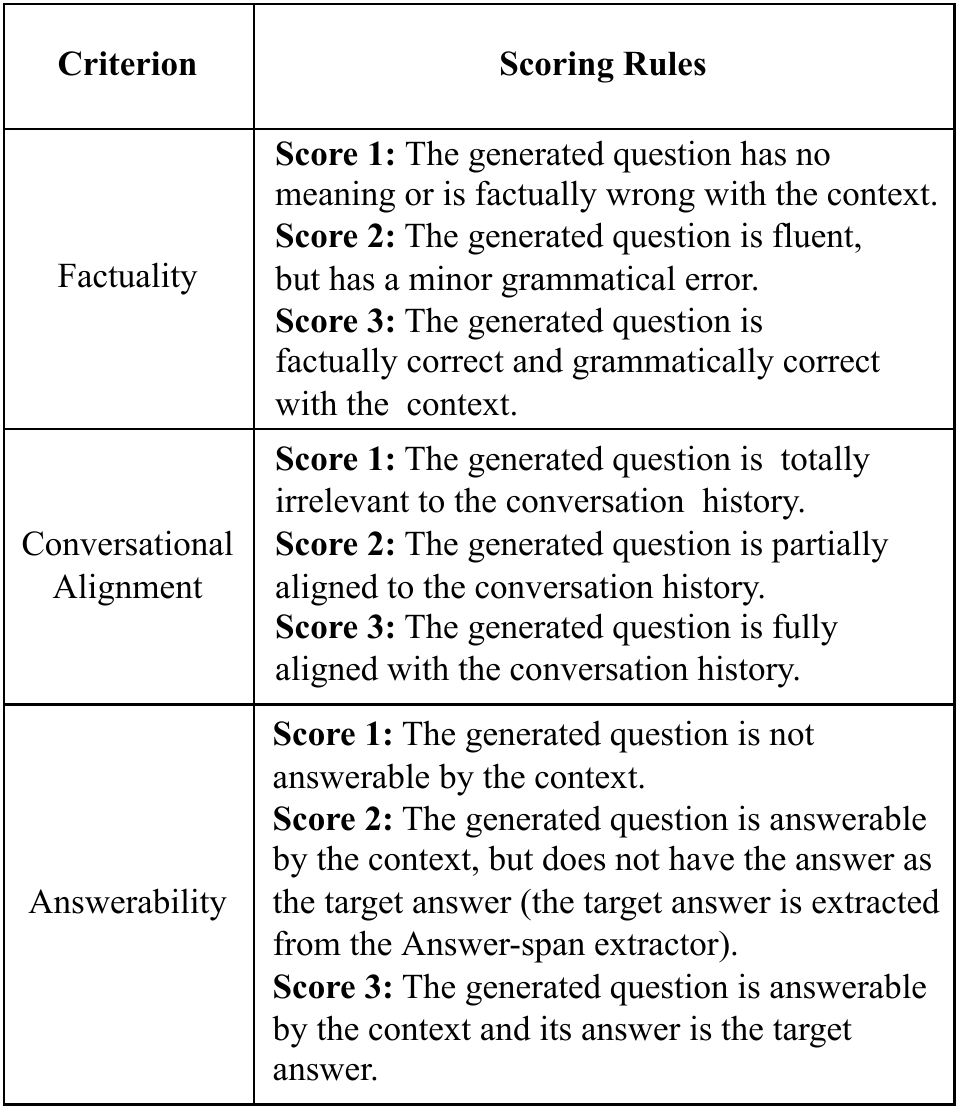}
\vspace{1mm}
{\begin{tabular}{p{15cm}}
\end{tabular}}
\vspace{-3mm}
\caption{\small Human Rating System
}
\vspace{-3mm}
\label{fig:human-rating-system}
\end{figure}

We describe how we instructed three annotators to point the generated questions based on three criteria: \emph{Factuality}, \emph{Conversational Alignment}, and \emph{Answerability}, as discussed in \Cref{sec:system-experiment}, in \Cref{fig:human-rating-system}. Following the discussion of \citet{cohs-cqg}, it is noted that in the answer-unaware setting, there is no target answer and rationale. However, since the what-to-ask module first seeks for one sentence in the context as the rationale and extracts the target answer from it, we still have the target answer and rationale. We thus define the \emph{Score 2, 3} based on the retrieved rationale and extracted target answer.

\subsection{Extended Case Studies}
\label{appendix:extended-case-studies}
We present additional samples generated by SG-CQG in \Cref{tab:extended-case-studies}. \change{In these samples, the first turn of the conversations is input to the model, and it generates their following turns.}

\begin{table*}[t!]
\centering
\small 
\resizebox{\linewidth}{!}
{
\begin{tabular}{p{7cm}p{6cm}p{1.5cm}}
\toprule
\textbf{Context} & \textbf{Generated Conversation} & \textbf{Rationales} \\
\midrule
1. Once there was a beautiful fish named \textcolor{yellow}{Asta}. \newline
2. Asta lived in \textcolor{blue}{the ocean}. \newline 
3. There were lots of other fish in the ocean where Asta lived. \newline
4. They played all day long. \newline
5. One day, a bottle floated by over the heads of \textcolor{orange}{Asta and his friends}. \newline
6. They looked up and saw the \textcolor{green}{bottle}. \newline
... \newline
10. "It looks like a \textcolor{brown}{bird's belly}," said Asta. \newline
... \newline
14. \textcolor{pear}{They wanted to see what was inside}. \newline 
... \newline 
18. The note was written in \textcolor{violet}{orange crayon} on \textcolor{yellow}{white paper}. \newline
... \newline 
26. This note is from a little girl. \newline
27. \textcolor{ruddybrown}{She wants to be your friend}. \newline 
... \newline 
&
Q1: What was the fish named? \newline
A1: \textcolor{yellow}{Asta} \newline
Q2: In what body of water? \newline
A2: \textcolor{blue}{The ocean} \newline
Q3: Who was in it? \newline
A3: \textcolor{orange}{Asta and his friends} \newline
Q4: What did they find? \newline
A4: \textcolor{green}{A bottle} \newline
Q5: Why did they discover it? \newline
A5: Because \textcolor{pear}{they wanted to see what was inside} \newline
Q6: What part of the bottle was it? \newline
A6: It was a \textcolor{brown}{bird's belly} \newline
Q7: What was the note written in? \newline
A7: \textcolor{violet}{Orange crayon} \newline
Q8: On what? \newline
A8: \textcolor{yellow}{White paper} \newline
Q9: What did the little girl want? \newline
A9: \textcolor{ruddybrown}{She wants to be your friend} \newline
&   
1,\newline 2,\newline 5,\newline 6,\newline 14,\newline 10,\newline 19,\newline 19,\newline 27 \\[-1em]
\midrule

1. This is the story of \textcolor{yellow}{a young girl and her \textcolor{blue}{dog}}. \newline
2. The young girl and her dog set out \textcolor{green}{a trip into the woods one day}. \newline
3. Upon entering the woods the girl and her dog found that the woods were \textcolor{brown}{dark and cold}. \newline
4. The girl was a little \textcolor{springgreen}{scared} and was thinking of turning back, but yet they went on. \newline
5. The girl's dog was acting very interested in what was in the bushes up ahead. \newline
6. To both the girl and the dog's surprise, there was \textcolor{orange}{a small brown bear} resting in the bushes. \newline
7. The bear was not surprised and did not seem at all interested in the girl and her dog. \newline
8. The bear looked up at the girl and it was almost as if he was \textcolor{springgreen}{smiling} at her. \newline
... \newline
10. The girl and the dog \textcolor{violet}{kept walking} and finally made it out of the woods. \newline 
... \newline

&
Q1: What is the story of? \newline
A1: \textcolor{yellow}{A young girl and her dog} \newline
Q2: What was the weather like? \newline
A2: \textcolor{brown}{Dark and cold} \newline
Q3: What did they go on? \newline
A3: \textcolor{green}{A trip into the woods} \newline
Q4: What kind of animal did they find? \newline
A4: \textcolor{orange}{A small brown bear} \newline
Q5: How did it make them feel? \newline
A5: \textcolor{springgreen}{Scared} \newline
Q6: How did they get out? \newline
A6: \textcolor{violet}{Kept walking} \newline
Q7: Did they have a \textcolor{blue}{dog}? \newline
A7: Yes \newline
Q8: How did the bear at her? \newline
A8: \textcolor{springgreen}{Smiling} \newline
&   
1,\newline 3,\newline 2,\newline 6,\newline 4,\newline 10, \newline 1, \newline 8 \\[-2em]

\bottomrule
\end{tabular}
}
\vspace{1mm}
\caption{\small{Additional sample conversations generated by our model SG-CQG. The first turn of both conversations is given to the model.}
}
\vspace{-5mm}
\label{tab:extended-case-studies}
\end{table*}


\subsection{Supplement: Why Do Control Signals Work?}
\label{appendix:how-do-control-signal-work}

In this section, we supplement the experimental results of the experiments with the \emph{control\_signal}. The results are presented in \Cref{tab:question-yes-no-generation-performance}, and the discussions are in \Cref{ssec:how-do-control-signals-work}.

\begin{table*}[htbp]
\centering
\scalebox{0.85}{

\makebox[\textwidth][c]{
    \begin{tabular}{l|ccc|ccc}
\toprule
         & \multicolumn{3}{c|}{ROUGE-L (boolean/span-based)} & 
         \multicolumn{3}{c}{BERTScore (boolean/span-based)} \\
\midrule
Model & Precision (\%) & Recall (\%) & F1 (\%) & Precision (\%) & Recall (\%) & F1 (\%) \\ 
\hline
\emph{With \texttt{Y/N}} & \textbf{38.70}/51.81 & \textbf{38.97}/53.06 & \textbf{37.73}/50.65 & \textbf{93.12}/\textbf{93.66} & \textbf{92.96}/\textbf{93.90} & \textbf{93.03}/\textbf{93.77} \\
\emph{W/o \texttt{Y/N}} & 35.92/51.81 & 35.49/53.07 & 34.59/50.64 & 92.70/\textbf{93.66} & 92.47/\textbf{93.90} & 92.57/\textbf{93.77} \\
\emph{W/o \texttt{Y/N} + $control\_signal$} & 37.56/\textbf{51.86} & 37.18/\textbf{53.09} & 36.22/\textbf{50.68} & 92.96/\textbf{93.66} & 92.74/\textbf{93.90} & 92.84/\textbf{93.77} \\
\bottomrule
\end{tabular}
}
}
\vspace{3mm}
\caption{Performance of T5 model in different settings. \emph{\texttt{Y/N}} denotes \emph{\texttt{Yes/No}} keyword.}

\label{tab:question-yes-no-generation-performance}
\vspace{-7mm}
\end{table*} 

\end{document}